\documentclass[11pt]{article}

\usepackage{acl2002}
\usepackage{times}
\usepackage{psfig}
\usepackage{amsmath}
\usepackage{latexsym}

\newcommand{\pinch}{\vspace{-.08in}}
\newcommand{\pinchh}{\vspace{-.05in}}

\title{A Noisy-Channel Model for Document Compression}
\author{
\begin{tabular}[t]{c}
Hal Daum\'e III and Daniel Marcu\\
Information Sciences Institute\\
University of Southern California\\
4676 Admiralty Way, Suite 1001\\
Marina del Rey, CA 90292\\
{\tt \{hdaume,marcu\}@isi.edu}\\
\end{tabular}
}

\date{}

\newcommand{\fto}{\rightarrow}
\renewcommand{\t}[1]{\textnormal{\small #1}}

\newcommand{\eqtext}[2]{
                       \vspace*{-3mm}
                       \begin{equation}
                       \mbox{ 
                       \begin{minipage}[t]{6.8cm}
                       \begin{small}  
                       {#1}
                       \end{small}
                       \end{minipage}
                       }
                       \label{#2}
                       \end{equation}
                       \vspace*{-1mm}
                       }

\begin{document}
\maketitle

\begin{abstract}
  We present a document compression system that uses a hierarchical
  noisy-channel model of text production.  Our compression system
  first automatically derives the syntactic structure of each sentence
  and the overall discourse structure of the text given as input.  The
  system then uses a statistical hierarchical model of text production
  in order to drop non-important syntactic and discourse constituents
  so as to generate coherent, grammatical document compressions of
  arbitrary length.  The system outperforms both a baseline and a
  sentence-based compression system that operates by simplifying
  sequentially all sentences in a text.  Our results support
  the claim that discourse knowledge plays an important
  role in document summarization.  
\end{abstract}

\section{Introduction} \label{intro} \pinch

Single document summarization systems proposed to date fall within one
of the following three classes:
\begin{description} \pinch
\item[Extractive summarizers] simply select and present to the user
  the most important sentences in a text ---
  see~\cite{mani-maybury-book99,marcu-book00,mani-book01} for
  comprehensive overviews of the methods and algorithms used to
  accomplish this.\pinch
\item[Headline generators] are noisy-channel probabilistic systems
  that are trained on large corpora of $\langle$\textit{Headline,
    Text}$\rangle$
  pairs~\cite{bankoetal00,berger-mittal00}.
  These systems produce short sequences of words that are indicative
  of the content of the text given as input. \pinch
\item[Sentence simplification
  systems]~\cite{chandrasekaretal96,mahesh97,carrolletal98,grefenstette98,jing00,knight-marcu00}
  are capable of compressing long sentences by deleting unimportant
  words and phrases.
\end{description}\pinch

Extraction-based summarizers often produce outputs that contain
non-important sentence fragments.  For example, the hypothetical
extractive summary of Text~\eqref{text1}, which is shown in
Table~\ref{sample-summaries}, can be compacted further by deleting the
clause ``which is already almost enough to win''.  Headline-based
summaries, such as that shown in Table~\ref{sample-summaries}, are
usually indicative of a text's content but not informative,
grammatical, or coherent.  By repeatedly applying a
sentence-simplification algorithm one sentence at a time, one can
compress a text; yet, the outputs generated in this way are likely to
be incoherent and to contain unimportant information.  When
summarizing text, some sentences should be dropped altogether.

\begin{table*}[t]
\leavevmode
\centering
\begin{small}

\begin{tabular}{|l|l|c|c|c|} \hline
Type of & Hypothetical output & Output & Output is & Output is \\
Summarizer &  & contains \textit{only} & coherent &
grammatical \\ 
& & important info & & \\ \hline\hline

Extractive & John Doe has already secured the vote of most  & & & $\surd$\\
summarizer & democrats in his constituency, which is already & & & \\
& almost enough to win.  But without the support & & & \\
& of the governer, he is still on shaky ground. & & &\\ \hline

Headline & mayor vote constituency governer & $\surd$ & & \\ 
generator & & & & \\ \hline

Sentence & The mayor is now looking for re-election.  John Doe & & & $\surd$ \\
simplifier & has already secured the vote of most democrats  & & & \\
               & in his constituency. He is still on shaky ground.  & & & \\\hline\hline

Document &  John Doe has secured the vote of most democrats. &
$\surd$ & $\surd$ & $\surd$ \\
compressor & But he is still on shaky ground. & & &\\\hline
\end{tabular}

\end{small}
\caption{Hypothetical outputs generated by various types of summarizers.}
\label{sample-summaries}
\end{table*}

Ideally, we would like to build systems that have the strengths of all
these three classes of approaches.  The ``Document Compression'' entry
in Table~\ref{sample-summaries} shows a grammatical, coherent summary
of Text~\eqref{text1}, which was generated by a hypothetical document
compression system that preserves the most important information in a
text while deleting sentences, phrases, and words that are subsidiary
to the main message of the text.  Obviously, generating coherent,
grammatical summaries such as that produced by the hypothetical
document compression system in Table~\ref{sample-summaries} is not
trivial because of many conflicting goals\footnote{A number of other
systems use the outputs of extractive summarizers and repair them to
improve coherence \cite{duc01,duc02}.  Unfortunately, none of these
seems flexible enough to produce in one shot good summaries that are
simultaneously coherent and grammatical.}.  The deletion of certain
sentences may result in incoherence and information loss.  The
deletion of certain words and phrases may also lead to
ungrammaticality and information loss.

\pinchh
\eqtext{The mayor is now looking for re-election.  John Doe has already
  secured the vote of most democrats in his constituency, which is
  already almost enough to win.  But without the support of the
  governer, he is still on shaky grounds.}{text1}
\pinch

In this paper, we present a document compression system that uses
hierarchical models of discourse and syntax in order to simultaneously
manage all these conflicting goals.  Our compression system first
automatically derives the syntactic structure of each sentence and the
overall discourse structure of the text given as input.  The system
then uses a statistical hierarchical model of text production in order
to drop non-important syntactic and discourse units so as to generate
coherent, grammatical document compressions of arbitrary length.  The
system outperforms both a baseline and a sentence-based compression
system that operates by simplifying sequentially all sentences in a
text.
\pinch
\section{Document Compression} \label{doc_comp} \pinch

The document compression task is conceptually simple.  Given a
document $D = \langle w_1 w_2 \dots w_n \rangle$, our goal is to
produce a new document $D'$ by ``dropping'' words $w_i$ from $D$.  In
order to achieve this goal, we extent the noisy-channel model proposed by
Knight \& Marcu \shortcite{knight-marcu00}.  Their system compressed
sentences by dropping syntactic constituents, but could be applied to
entire documents only on a sentence-by-sentence basis.  As discussed
in Section~\ref{intro}, this is not adequate because the resulting
summary may contain many compressed sentences that are irrelevant.
In order to extend Knight \& Marcu's approach beyond the sentence
level, we need to ``glue'' sentences together in a tree
structure similar to that used at the sentence level.  Rhetorical
Structure Theory (RST) \cite{mann88} provides us this
``glue.''

\begin{figure*}[t]
\center\mbox{\psfig{figure=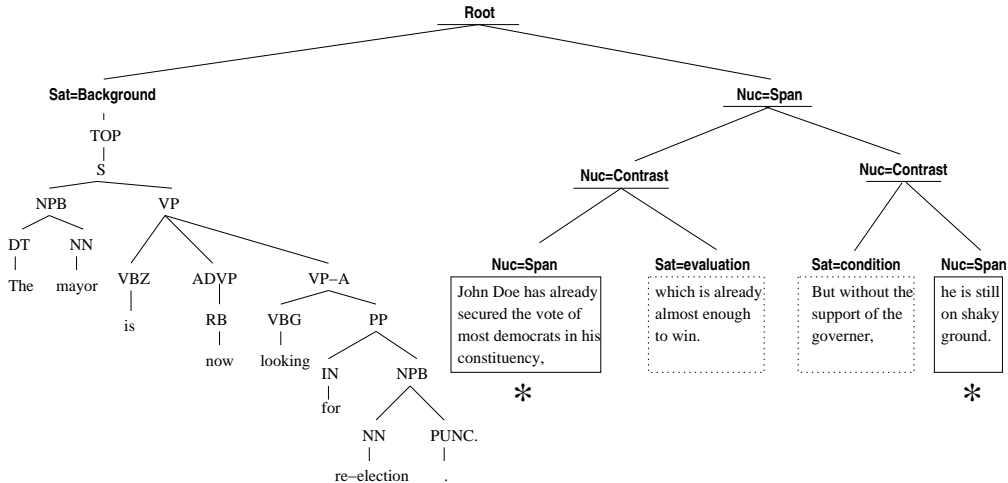,height=2.5in}}
\caption{The discourse (full)/syntax (partial) tree for Text~\eqref{text1}.}
\label{dstree}
\end{figure*}

The tree in Figure~\ref{dstree} depicts the RST structure of
Text~\eqref{text1}.  In RST, discourse structures are non-binary trees
whose leaves correspond to {\em elementary discourse units (EDUs)},
and whose internal nodes correspond to contiguous text spans.  Each
internal node in an RST tree is characterized by a {\em rhetorical
relation}.  For example, the first sentence in Text~\eqref{text1}
provides \textsc{\small background} information for interpreting the
information in sentences 2 and 3, which are in a \textsc{\small
Contrast} relation (see Figure~\ref{dstree}).  Each relation holds
between two adjacent non-overlapping text spans called \textsc{\small
nucleus} and \textsc{\small satellite}.  (There are a few exceptions
to this rule: some relations, such as \textsc{\small list} and
\textsc{\small contrast}, are multinuclear.)  The distinction between
nuclei and satellites comes from the empirical observation that the
nucleus expresses what is more essential to the writer's purpose than
the satellite.

Our system is able to analyze both the discourse structure of a
document and the syntactic structure of each of its sentences or EDUs.
It then compresses the document by dropping either
syntactic or discourse constituents.

\pinch
\section{A Noisy-Channel Model} \label{model}
\pinch 

For a given document $D$, we want to find the summary text $S$ that
maximizes $P(S|D)$.  Using Bayes rule, we flip this so we end up
maximizing ${P(D|S)P(S)}$.  Thus, we are left with modelling two
probability distributions: $P(D|S)$, the probability of a document $D$
given a summary $S$, and $P(S)$, the probability of a summary.  We
assume that we are given the discourse structure of each document and
the syntactic structures of each of its EDUs.

The intuitive way of thinking about this application of Bayes rule,
reffered to as the noisy-channel model, is that we start with a
summary $S$ and add ``noise'' to it, yielding a longer document $D$.
The noise added in our model consists of words, phrases and discourse
units.

For instance, given the document ``John Doe has secured the vote of
most democrats.'' we could add words to it (namely the word
``already'') to generate ``John Doe has already secured the vote of
most democrats.''  We could also choose to add an entire syntactic
constituent, for instance a prepositional phrase, to generate ``John
Doe has secured the vote of most democrats \emph{in his
constituency}.''  These are both examples of sentence expansion as
used previously by Knight \& Marcu \shortcite{knight-marcu00}.

Our system, however, also has the ability to expand on a core message
by adding discourse constituents.  For instance, it could decide to add
another discourse constituent to the original summary ``John Doe has
secured the vote of most democrats'' by \textsc{\small contrast}ing
the information in the summary with the uncertainty regarding the
support of the governor, thus yielding the text: ``John Doe has
secured the vote of most democrats.  \emph{But without the support of
the governor, he is still on shaky ground}.''

As in any noisy-channel application, there are three parts that we
have to account for if we are to build a complete document compression
system: the channel model, the source model and the decoder.  We
describe each of these below.
\begin{description}\pinchh
\item[The source model] assigns to a string the probability $P(S)$,
the probability that the summary $S$ is good English.  Ideally, the
source model should disfavor ungrammatical sentences and documents
containing incoherently juxtaposed sentences.
\pinch
\item[The channel model] assigns to any
document/summary pair a probability $P(D|S)$.  This models the extent to which $D$ is
a good expansion of $S$.  For instance, if $S$ is ``The mayor is now
looking for re-election.'', $D_1$ is ``The mayor is now looking for
re-election.  \emph{He has to secure the vote of the democrats.}'' and
$D_2$ is ``The major is now looking for re-election.  \emph{Sharks
have sharp teeth.}'', we expect $P(D_1|S)$ to be higher than
$P(D_2|S)$ because $D_1$ expands on $S$ by elaboration, while
$D_2$ shifts to a different topic, yielding an incoherent text.
\pinch
\item[The decoder] searches through all possible summaries of a
  document $D$ for the summary $S$ that maximizes the posterior
  probability $P(D|S)P(S)$.
\end{description}
Each of these parts is described below.

\subsection{Source model}

The job of the source model is to assign a score $P(S)$ to a
compression independent of the original document.  That is, the source
model should measure how good English a summary is (independent of
whether it is a good compression or not).  Currently, we use a bigram
measure of quality (trigram scores were also tested but failed to make
a difference), combined with non-lexicalized context-free syntactic
probabilities and context-free discourse probabilities, giving $P(S) =
P_{bigram}(S) * P_{PCFG}(S) * P_{DPCFG}(S)$.  It would be better to
use a lexicalized context free grammar, but that was not possible
given the decoder used.

\subsection{Channel model} \label{channel}

The channel model is allowed to add syntactic constituents (through a
stochastic operation called \emph{constituent-expand}) or discourse
units (through another stochastic operation called \emph{EDU-expand}).
Both of these operations are performed on a combined discourse/syntax
tree called the DS-tree.  The DS-tree for Text~\eqref{text1} is shown
in Figure~\ref{dstree} for reference.

Suppose we start with the summary $S=$ ``The mayor is looking for
re-election.''  A constituent-expand operation could insert a
syntactic constituent, such as ``this year'' anywhere in the syntactic
tree of $S$.  A constituent-expand operation could also add single
words: for instance the word ``now'' could be added between ``is'' and
``looking,'' yielding $D=$ ``The mayor is \emph{now} looking for
re-election.''  The probability of inserting this word is based on the
syntactic structure of the node into which it's inserted. 

Knight and Marcu~\shortcite{knight-marcu00} describe in detail a
noisy-channel model that explains how short sentences can be expanded
into longer ones by inserting and expanding syntactic constituents
(and words). Since our \emph{constituent-expand} stochastic operation
simply reimplements Knight and Marcu's model, we do not focus on them
here.  We refer the reader to~\cite{knight-marcu00} for the details.

\begin{figure*}[t]
\center\mbox{\psfig{figure=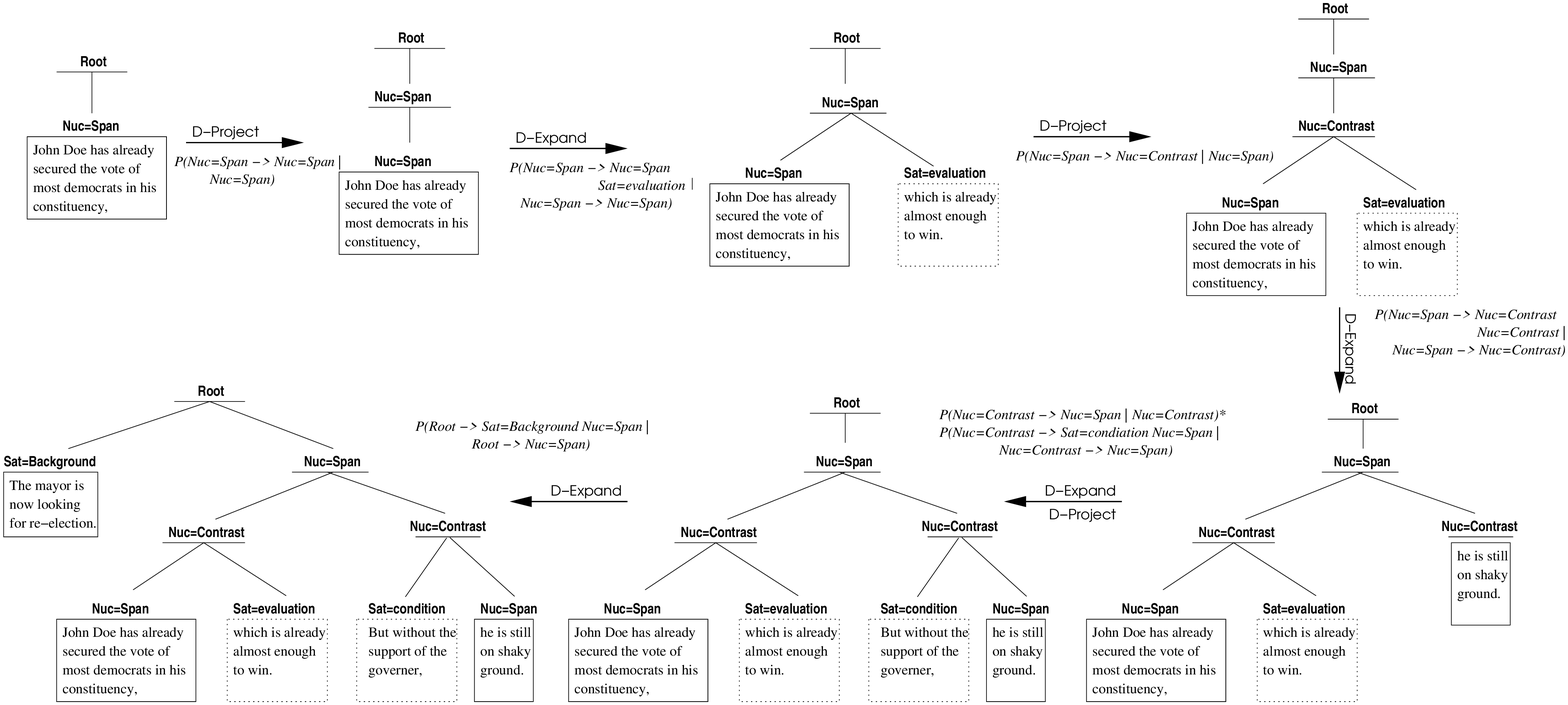,height=2.7in}}
\caption{A sequence of discourse expansions for
Text~\eqref{text1} (with probability factors).}
\label{example}
\end{figure*}

In addition to adding syntactic constituents, our system is also able
to add discourse units.  Consider the summary $S=$ ``John Doe has
already secured the vote of most democrats in his consituency.''
Through a sequence of discourse expansions, we can expand upon this
summary to reach the original text.  A complete discourse expansion
process that would occur starting from this initial summary to
generate the original document is shown in Figure~\ref{example}.

In this figure, we can follow the sequence of steps required to
generate our original text, beginning with our summary $S$.  First,
through an operation \emph{D-Project} (``D'' for ``D''iscourse), we
increase the depth of the tree, adding an intermediate
\textsc{Nuc=Span} node.  This projection adds a factor of
$P(\t{Nuc=Span }\fto\t{ Nuc=Span}|\t{ Nuc=Span})$ to the probability
of this sequence of operations (as is shown under the arrow).

We are now able to perform the second operation, \emph{D-Expand}, with
which we expand on the core message contained in $S$ by adding a
satellite which evaluates the information presented in $S$.  This
expansion adds the probability of performing the expansion (called the
discourse expansion probabilities, $P_{DE}$.  An example discourse
expansion probability, written $P(\t{Nuc=Span }\fto\t{ Nuc=Span
Sat=Eval}|\t{ Nuc=Span }\fto\t{ Nuc=Span})$, reflects the probability
of adding an evaluation satellite onto a nuclear span).

The rest of Figure~\ref{example} shows some of the remaining steps to
produce the original document, each step labeled with the appropriate
probability factors.  Then, the probability of the entire expansion is
the product of all those listed probabilities combined with the
appropriate probabilities from the syntax side of things.  In order to
produce the final score $P(D|S)$ for a document/summary pair, we
multiply together each of the expansion probabilities in the path
leading from $S$ to $D$.

For estimating the parameters for the discourse models, we used an RST
corpus of 385 Wall Street Journal articles from the Penn Treebank,
which we obtained from LDC.  The documents in the corpus range in size
from 31 to 2124 words, with an average of 458 words per document.
Each document is paired with a discourse structure that was manually
built in the style of RST.  (See~\cite{marcu-rst-corpus01} for details
concerning the corpus and the annotation process.)  From this corpus,
we were able to estimate parameters for a discourse PCFG using
standard maximum likelihood methods.

Furthermore, 150 document from the same corpus are paired with
extractive summaries on the EDU level.  Human annotators were asked
which EDUs were most important; suppose in the example DS-tree
(Figure~\ref{dstree}) the annotators marked the second and fifth EDUs
(the starred ones).  These stars are propagated up, so that any
discourse unit that has a descendent considered important is also
considered important.  From these annotations, we could deduce that,
to compress a \textsc{\small Nuc=Contrast} that has two children,
\textsc{\small Nuc=Span} and \textsc{\small Sat=evaluation}, we can
drop the evaluation satellite.  Similarly, we can compress a
\textsc{\small Nuc=Contrast} that has two children, \textsc{\small
Sat=condition} and \textsc{\small Nuc=Span} by dropping the first
discourse constituent.  Finally, we can compress the \textsc{\small
Root} deriving into \textsc{\small Sat=Background Nuc=Span} by
dropping the \textsc{\small Sat=Background} constituent.  We keep
counts of each of these examples and, once collected, we normalize
them to get the discourse expansion probabilities.

\subsection{Decoder} \label{decoder}

The goal of the decoder is to combine $P(S)$ with $P(D|S)$ to get
$P(S|D)$.  There are a vast number of potential compressions of a
large DS-tree, but we can efficiently pack them into a shared-forest
structure, as described in detail by Knight \& Marcu
\shortcite{knight-marcu00}.  Each entry in the shared-forest structure
has three associated probabilities, one from the source syntax PCFG,
one from the source discourse PCFG and one from the expansion-template
probabilities described in Section~\ref{channel}.  Once we have
generated a forest representing all possible compressions of the
original document, we want to extract the best (or the $n$-best)
trees, taking into account both the expansion probabilities of the
channel model and the bigram and syntax and discourse PCFG probabilities of the source model.
Thankfully, such a generic extractor has already been built
\cite{langkilde00}.  For our purposes, the extractor selects the trees
with the best combination of LM and expansion scores after
performing an exhaustive search over all possible summaries.  It
returns a list of such trees, one for each possible length.

\section{System} \label{system}

\begin{figure}[!h]
\center\mbox{\psfig{figure=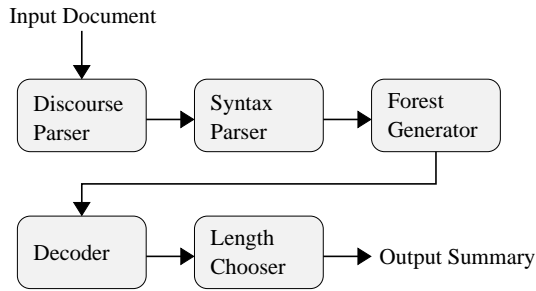,height=1.5in}}
\caption{The pipeline of system components.}
\label{systemfig}
\end{figure}

The system developed works in a pipelined fashion as shown in
Figure~\ref{systemfig}.  The first step along the pipeline is to
generate the discourse structure.  To do this, we use the
decision-based \emph{discourse parser} described by Marcu
\shortcite{marcu-book00}\footnote{The discourse parser achieves an
f-score of $38.2$ for EDU identification, $50.0$ for identifying
hierarchical spans, $39.9$ for nuclearity identification and $23.4$
for relation tagging.}.  Once we have the discourse structure, we send
each EDU off to a \emph{syntactic parser} \cite{collins-acl97}.  The
syntax trees of the EDUs are then merged with the discourse tree in
the \emph{forest generator} to create a DS-tree similar to that shown
in Figure~\ref{dstree}.  From this DS-tree we generate a forest that
subsumes all possible compressions.  This forest is then passed on to
the forest ranking system which is used as \emph{decoder}
\cite{langkilde00}.  The decoder gives us a list of possible
compressions, for each possible length.  Example compressions of
Text~\eqref{text1} are shown in Figure~\ref{compressions} together
with their respective log-probabilities.

\begin{figure*}[t]
\mbox{\small
\begin{tabular}{rlp{5in}}
\textbf{len} & \textbf{log prob} & \textbf{best compression} \\
$8$ & $-118.9060$ &  \emph{Mayor is now looking which is enough.} \\
$13$ & $-137.1010$ &  The mayor is now looking which is already almost enough to win.  \\
$16$ & $-147.5970$ &  The mayor is now looking but without support, he is still on shaky ground.  \\
$18$ & $-160.4310$ &  Mayor is now looking but without the support of governer, he is still on shaky ground.  \\
$22$ & $-176.1990$ &  The mayor is now looking for re-election but without the support of the governer, he is still on shaky ground.  \\
$28$ & $-239.9490$ &  The mayor is now looking which is already almost enough to win.  But without the support of the governer, he is still on shaky ground.  \\
\end{tabular}
}
\caption{Possible compressions for Text~\eqref{text1}.}
\label{compressions}
\end{figure*}

In order to choose the ``best'' compression at any possible length, we
cannot rely only on the log-probabilities, lest the system always
choose the shortest possible compression.  In order to compensate for
this, we normalize by length.  However, in practice, simply dividing
the log-probability by the length of the compression is insufficient
for longer documents.  Experimentally, we found a reasonable metric
was to, for a compression of length $n$, divide each log-probability
by $n^{1.2}$.  This was the job of the \emph{length chooser} from
Figure~\ref{systemfig}, and enabled us to choose a single compression
for each document, which was used for evaluation.  (In
Figure~\ref{compressions}, the compression chosen by the length
selector is italicized and was the shortest one\footnote{This tends to
be the case for very short documents, as the compressions never get
sufficiently long for the length normalization to have an effect.}.)

\section{Results} \label{results}

For testing, we began with two sets of data.  The first set is drawn
from the Wall Street Journal (WSJ) portion of the Penn Treebank and
consists of $16$ documents, each containing between $41$ and $87$
words.  The second set is drawn from a collection of student
compositions and consists of $5$ documents, each containing between
$64$ and $91$ words. We call this set the MITRE
corpus~\cite{hirschman99deep}.  We would liked to have run evaluations
on longer documents. Unfortunately, the forests generated even for
relatively small documents are huge. Because there are an exponential
number of summaries that can be generated for any given
text\footnote{In theory, a text of $n$ words has $2^n$ possible
compressions.}, the decoder runs out of memory for longer documents;
therefore, we selected shorter subtexts from the original documents.

We used both the WSJ and Mitre data for evaluation because we wanted
to see whether the performance of our system varies with text genre.
The Mitre data consists mostly of short sentences (average document
length from Mitre is $6$ sentences), quite in constrast to the
typically long sentences in the Wall Street Journal articles (average
document length from WSJ is $3.25$ sentences).

For purpose of comparison, the Mitre data was compressed using five
systems: 
\begin{description}
\pinchh
\item [Random:] Drops random words (each word has a 50\% chance of
being dropped (baseline).
\pinchh
\item [Hand:] Hand compressions done by a human.
\pinchh
\item [Concat:] Each sentence is compressed individually; the results
are concatenated together, using Knight \& Marcu's
\shortcite{knight-marcu00} system here for comparison.  \pinchh
\item [EDU:] The system described in this paper.
\pinchh
\item [Sent:] Because syntactic parsers tend not to work well
parsing just clauses, this system merges together leaves in the
discourse tree which are in the same sentence, and then proceeds as
described in this paper.
\pinchh
\end{description}

The Wall Street Journal data was evaluated on the above five systems
as well as two additions.  Since the correct discourse trees were
known for these data, we thought it wise to test the systems using
these human-built discourse trees, instead of the automatically
derived ones.  The additionall two systems were:
\begin{description}
\pinchh
\item [PD-EDU:] Same as {\bf EDU} except using the perfect discourse
trees, available from the RST corpus \cite{marcu-rst-corpus01}.
\pinchh
\item [PD-Sent:] The same as {\bf Sent} except using the perfect
discourse trees.
\pinchh
\end{description}
Six human evaluators rated the systems according to three metrics.
The first two, presented together to the evaluators, were
grammaticality and coherence; the third, presented separately, was
summary quality.  Grammaticality was a judgment of how good the
English of the compressions were; coherence included how well the
compression flowed (for instance, anaphors lacking an antecedent would
lower coherence).  Summary quality, on the other hand, was a judgment
of how well the compression retained the meaning of the original
document.  Each measure was rated on a scale from $1$ (worst) to $5$
(best).

We can draw several conclusions from the evaluation results shown in
Table~\ref{results_table} along with average compression rate
(\emph{Cmp}, the length of the compressed document divided by the
original length).\footnote{We did not run the system on the MITRE data
with perfect discourse trees because we did not have hand-built
discourse trees for this corpus.}  First, it is clear that genre
influences the results.  Because the Mitre data contained mostly short
sentences, the syntax and discourse parsers made fewer errors, which
allowed for better compressions to be generated. For the Mitre corpus,
compressions obtained starting from discourse trees built above the
sentence level were better than compressions obtained starting from
discourse trees built above the EDU level. For the WSJ corpus,
compression obtained starting from discourse trees built above the
sentence level were more grammatical, but less coherent than
compressions obtained starting from discourse trees built above the
EDU level. Choosing the manner in which the discourse and syntactic
representations of texts are mixed should be influenced by the genre
of the texts one is interested to compress.

\begin{table}[tb]
\center\small
\mbox{
\begin{tabular}{|@{ }r@{ }||c@{ }c@{ }c@{ }c@{ }||c@{ }c@{ }c@{ }c@{ }|}
\hline
        & \multicolumn{4}{c||}{WSJ}     & \multicolumn{4}{c|}{Mitre} \\
        & Cmp & Grm & Coh & Qual              & Cmp & Grm & Coh & Qual        \\
\hline
Random  & 0.51 & 1.60 & 1.58 & 2.13    & 0.47 & 1.43 & 1.77 & 1.80 \\
Concat  & 0.44 & 3.30 & 2.98 & 2.70    & 0.42 & 2.87 & 2.50 & 2.08 \\
EDU     & 0.49 & 3.36 & 3.33 & 3.03    & 0.47 & 3.40 & 3.30 & 2.60 \\
Sent    & 0.47 & 3.45 & 3.16 & 2.88    & 0.44 & 4.27 & 3.63 & 3.36 \\
PD-EDU  & 0.47 & 3.61 & 3.23 & 2.95    & \multicolumn{4}{c|}{}    \\
PD-Sent & 0.48 & 3.96 & 3.65 & 2.84    & \multicolumn{4}{c|}{}    \\
Hand    & 0.59 & 4.65 & 4.48 & 4.53    & 0.46 & 4.97 & 4.80 & 4.52  \\
\hline
\end{tabular}
}
\caption{Evaluation Results}
\label{results_table}
\end{table}

The compressions obtained starting from perfectly derived discourse
trees indicate that perfect discourse structures help greatly in
improving coherence and grammaticality of generated summaries. It was
surprising to see that the summary quality was affected negatively by
the use of perfect discourse structures (although not statistically
significant). We believe this happened because the text fragments we
summarized were extracted from longer documents.  It is likely that
had the discourse structures been built specifically for these short
text snippets, they would have been different.  Moreover, there was no
component designed to handle cohesion; thus it is to be
expected that many compressions would contain dangling references.

Overall, all our systems outperformed both the Random baseline and the
Concat systems, which empirically show that discourse has an important
role in document summarization. We performed $t$-tests on the results
and found that on the Wall Street Journal data, the differences in
score between the Concat and Sent systems for grammaticality and
coherence were statistically significant at the 95\% level, but the
difference in score for summary quality was not.  For the Mitre data,
the differences in score between the Concat and Sent systems for
grammaticality and summary quality were statistically significant at
the 95\% level, but the difference in score for coherence was not. The
score differences for grammaticality, coherence, and summary quality
between our systems and the baselines were statistically significant
at the 95\% level.

The results in Table~\ref{results_table}, which can be also assessed
by inspecting the compressions in Figure~\ref{compressions} show that,
in spite of our success, we are still far away from human performance
levels. An error that our system makes often is that of dropping
complements that cannot be dropped, such as the phrase ``for
re-election'', which is the complement of ``is looking''. We are
currently experimenting with lexicalized models of syntax that would
prevent our compression system from dropping required verb arguments.
We also consider methods for scaling up the decoder to handling
documents of more realistic length.

\section*{Acknoledgements}

This work was partially supported by DARPA-ITO grant N66001-00-1-9814,
NSF grant IIS-0097846, and a USC Dean Fellowship to Hal Daume III.
Thanks to Kevin Knight for discussions related to the project.

%\section{Conclusions and Future Work} \label{conclusion}

%Referring back to the compressions generated for Text~\eqref{text1} in
%Figure~\ref{compressions}, there are a few mistakes the system makes
%which could be avoided.  For example, the system has learned that it
%can often drop prepositional phrases.  Unfortunately, since the rules
%used by the system are not lexicalized, the system cannot know that it
%should not drop the prepositional phrase ``for re-election'', which is
%the complement of ``is looking.''  This is a general problem with the
%system: it often drops complements of transitive verbs.  Problems like
%these would be resolved by using a syntax-based language model (in
%fact, simply by reranking the the n-best list from the decoder using a
%syntax-based scorer, the chosen compression becomes ``The mayor is now
%looking for re-election which is enough to win. But without the
%support, he is still on ground.'', which is a clear improvement).

\bibliographystyle{acl}
\bibliography{bibfile}

\end{document}